\documentclass[sigplan,nonacm]{acmart}


\usepackage{fancyhdr}

\usepackage{etoolbox}

\usepackage{siunitx}

\usepackage{color, colortbl}
\usepackage{url}            
\usepackage{booktabs}       
\usepackage{amsfonts}       
\usepackage{nicefrac}       
\usepackage{xcolor}         
\usepackage{xspace}
\usepackage{wrapfig}
\usepackage{graphicx}
\usepackage{makecell}
\usepackage{subcaption}
\usepackage{multirow}
\usepackage{tabularx}
\usepackage{lipsum}
\usepackage{balance}

\usepackage{float}

\usepackage{adjustbox}
\usepackage[normalem]{ulem}

\usepackage{algpseudocode}
\setcitestyle{numbers,sort}

\usepackage{titlesec}
\usepackage[subtle]{savetrees}

\usepackage{stackengine}[2013-10-15]

\usepackage{amsmath}
\usepackage{comment}
\usepackage{xcolor}
\usepackage{xspace}
\usepackage{hyperref}
\usepackage{doi}

\usepackage{tikz}

\AtBeginShipoutFirst{%
  \begin{tikzpicture}[remember picture, overlay]
    \node[anchor=south, yshift=1cm] at (current page.south) {\Large Accepted for publication at ASPLOS 2025};
  \end{tikzpicture}%
}

\newcommand\todo[1]{\noindent{\color{orange} {\bf \fbox{TODO}} {\it#1}}}

\newcommand{\malian}[1]{{\color{red}[\textbf{\sc malian}: \textit{#1}]}}

\newcommand{\derrick}[1]{{\color{green}[\textbf{\sc derrick}: \textit{#1}]}}

\newcommand{\edit}[1]{{\color{black}{#1}}}
\newcommand{\editlow}[1]{{\color{black}{#1}}}

\renewcommand\sout[1]{}

\def\train{\textit{nq-train}\xspace}
\def\dev{\textit{nq-dev}\xspace}

\def\smalltable{\scriptsize}

\newcommand{\nonREML}{non-REML }

\def\re{RAG\xspace}
\def\knn{k-NN\xspace}
\def\enns{ENNS\xspace}
\def\anns{ANNS\xspace}
\def\k{k\xspace}

\def\candidate{indexed embedding\xspace}

\def\appt{\texttt{FiDT5}\xspace}
\def\appl{\texttt{Llama-70B}\xspace} 
\def\appm{\texttt{Llama-8B}\xspace} 
\def\querysp{Query Scratchpad\xspace}
\def\refp{reference point\xspace}

\title{Accelerating Retrieval-Augmented Generation}

\author{Derrick Quinn}
\affiliation{
\institution{Cornell University}
\city{Ithaca}
\state{NY}
\country{USA}}
\email{dq55@cornell.edu}

\author{Mohammad Nouri}
\affiliation{
\institution{Cornell University}
\city{Ithaca}
\state{NY}
\country{USA}}\email{mn636@cornell.edu}

\author{Neel Patel}
\affiliation{
\institution{Cornell University}
\city{Ithaca}
\state{NY}
\country{USA}}\email{nmp83@cornell.edu}

\author{John Salihu}
\affiliation{
\institution{University of Kansas}
\city{Lawrence}
\state{KS}
\country{USA}}\email{jsalihu@ku.edu}

\author{Alireza Salemi}
\affiliation{
\institution{University of Massachusetts Amherst}
\city{Amherst}
\state{MA}
\country{USA}}\email{asalemi@cs.umass.edu}

\author{Sukhan Lee}
\affiliation{
\institution{Samsung Electronics}
\city{Hwasung}
\country{Republic of Korea}}\email{sh1026.lee@samsung.com}

\author{Hamed Zamani}
\affiliation{
\institution{University of Massachusetts Amherst}
\city{Amherst}
\state{MA}
\country{USA}}\email{zamani@cs.umass.edu}

\author{Mohammad Alian}
\affiliation{
\institution{Cornell University}
\city{Ithaca}
\state{NY}
\country{USA}}\email{malian@cornell.edu}


\begin{document}

\thispagestyle{plain}

\pagestyle{fancy}

\def\BibTeX{{\rm B\kern-.05em{\sc i\kern-.025em b}\kern-.08em
    T\kern-.1667em\lower.7ex\hbox{E}\kern-.125emX}}

\begin{abstract}

An evolving solution to address hallucination and enhance accuracy in large language models (LLMs) is Retrieval-Augmented Generation (RAG), which involves augmenting LLMs with information retrieved from an external knowledge source, such as the web. This paper profiles several RAG execution pipelines and demystifies the complex interplay between their retrieval and generation phases. We demonstrate that while exact retrieval schemes are expensive, they can reduce inference time compared to approximate retrieval variants because an exact retrieval model can send a smaller but more accurate list of documents to the generative model while maintaining the same end-to-end accuracy. This observation motivates the acceleration of the exact nearest neighbor search for RAG. 

In this work, we design Intelligent Knowledge Store (IKS), a type-2 CXL device that implements a scale-out near-memory acceleration architecture with a novel cache-coherent interface between the host CPU and near-memory accelerators. IKS offers 13.4--27.9$\times$ faster exact nearest neighbor search over a 512GB vector database compared with executing the search on Intel Sapphire Rapids CPUs. This higher search performance translates to 1.7--26.3$\times$ lower end-to-end inference time for representative RAG applications. IKS is inherently a memory expander; its internal DRAM can be disaggregated and used for other applications running on the server to prevent DRAM -- which is the most expensive component in today's servers -- from being stranded.

\end{abstract}

\maketitle
\renewcommand{\shortauthors}{Derrick Quinn et al.}

\section{Introduction}
\label{sec:intro}

State-of-the-art natural language processing systems heavily rely on large language models (LLMs)--deep Transformer networks~\cite{transformer} with hundreds of millions of parameters. There is much evidence that information presented in the LLM training corpora is ``memorized'' in the LLM parameters, forming a parametric knowledge base that the model depends on for generating responses. A major challenge with parametric knowledge is its static nature; it cannot be updated unless the model undergoes retraining or fine-tuning, which is an extremely costly process. This creates a critical issue, especially when it comes to non-stationary domains where fresh content is constantly being produced~\cite{reml}. Besides, previous studies have indicated that LLMs exhibit limited memorization for less frequent entities~\cite{kandpal2022deduplicating}, are susceptible to hallucinations~\cite{shuster-etal-2021-retrieval-augmentation}, and may experience temporal degradation~\cite{kasai2022realtime}.

\begin{figure}
    \centering
    \includegraphics[width=0.9\linewidth]{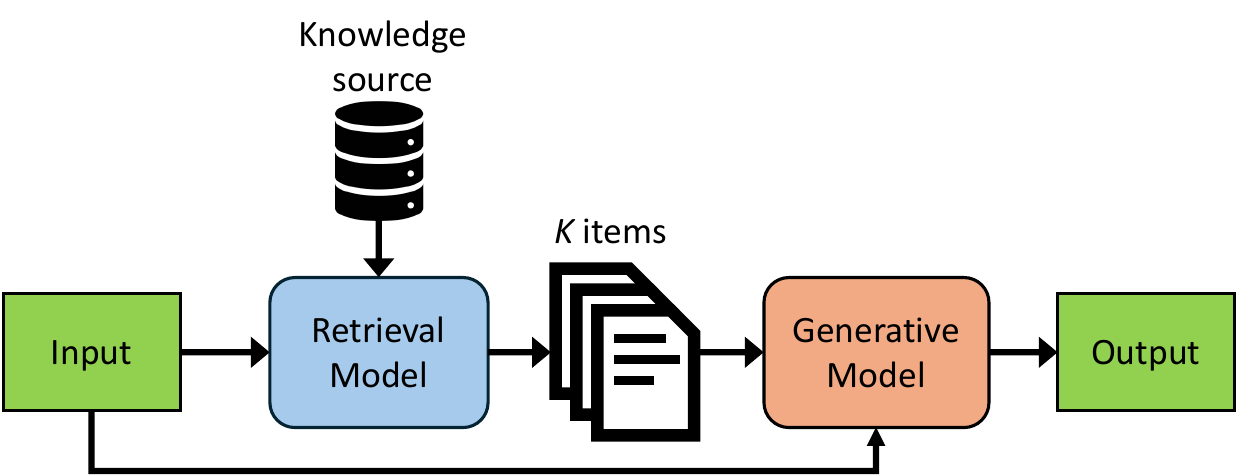}
    \caption{Overview of the Retrieval-Augmented Generation (RAG) pipeline.}
    \vspace{-10pt}
    \label{fig:rag-overview}
\end{figure}

To overcome the challenges presented by LLMs, a potential solution is to enhance them with non-parametric knowledge, where the LLM is augmented with information retrieved from a knowledge source (e.g., text documents). These approaches have recently gained considerable attention in the machine learning communities~\cite{10.5555/3495724.3496517,fid-light,liu2021retrievalaugmented,parvez-etal-2021-retrieval-augmented,shuster-etal-2021-retrieval-augmentation,salemi2023lamp,izacard-grave-2021-leveraging}, and have played key roles in some recent breakthrough applications in the tech industry, such as Google Gemini~\cite{geminiteam2024gemini}, Microsoft Copilot~\cite{azure-rag}, and OpenAI ChatGPT with Retrieval Plugins~\cite{gpt-plugins}.  Retrieval-Augmented Generation (RAG) is the term that is used to refer to systems that adopt this approach in the context of LLMs.

A RAG application includes two key components: a retrieval model and an LLM for text generation, called the generative model. When a query is received, the retrieval model searches for relevant items (e.g., documents) and the top retrieved items, together with the input, are sent to the generative model. Current state-of-the-art retrieval approaches use bi-encoder neural networks (called dense retrieval)~\cite{dpr} for learning optimal embedding for queries and documents. Each item is then encoded into a high-dimension vector (called embedding vectors) and stored in a vector database. Such approaches use K-nearest neighbor algorithms for retrieving the top ``K" items from the vector database.
Figure~\ref{fig:rag-overview} provides an overview of a RAG application. 

The accuracy of generated output in RAG hinges on the quality of the retrieved item list. Conducting an Exact Nearest Neighbors Search (\enns) to retrieve the precise top K relevant items involves scanning all the embedding vectors in the vector database, which is costly in today's memory bandwidth-limited systems. For example, in a RAG application with a 50GB (using a 16-bit floating point representation) vector database running on an Intel Xeon 4416+ with 8$\times$DDR5-4000 memory channels, and a generative model running on an \editlow{NVIDIA H100 GPU}, \enns takes up to 97\% of the end-to-end inference time (\S\ref{sec:profiling}).

One strategy to mitigate the retrieval cost is to employ Approximate Nearest Neighbor Search (\anns), and opt for a faster, but lower-quality search configuration. While lower-quality retrieval can improve search time, our extensive experiments on Question Answering applications demonstrate that a lower-quality search scheme should provide significantly more items to the language model in order to match the end-to-end RAG accuracy of \enns or a higher-quality, but slower \anns configuration. This virtually negates any benefits gained during the retrieval phase and even increases the end-to-end inference time.  

In this paper, we extensively profile the execution pipeline of RAG, demystifying the complex interplay between various hardware and software configurations in RAG applications~\cite{Zaharia-2024-Shift}. Motivated by the need for high-performance, low-cost, high-quality search and the limitations of current commodity systems, we contribute the Intelligent Knowledge Store (IKS), a cost-optimized, purpose-built CXL memory expander that functions as a high-performance, high-capacity vector database accelerator. IKS offloads memory-intensive dot-product operations in \enns to a distributed array of low-profile accelerators placed near LPDDR5X DRAM packages.


IKS implements a novel interface atop the CXL.cache protocol to seamlessly offload exact vector database search operations to near-memory accelerators. IKS is exposed as a memory expander that disaggregates its internal DRAM capacity and shares it with vector database applications and other co-running applications through CXL.mem and CXL.cache protocols. Instead of building a full-fledged vector database accelerator, IKS co-designs the hardware and software to implement a minimalist scale-out near-memory accelerator architecture. This design relies on software to map data into the internal IKS DRAM and scratchpads while performing the final top-K aggregation. 

In summary, we make the following contributions:
\begin{itemize}
    \item We demystify RAG by profiling its execution pipeline. We explore various hardware, system, and application-level configurations to assess the performance and accuracy of RAG.


    \item We demonstrate that RAG requires high-quality retrieval to perform effectively; nonetheless, current RAG applications are bottlenecked by a high-quality retrieval phase. 
    
    \item We introduce Intelligent Knowledge Store (IKS), which is a specialized CXL-based memory expander equipped with low-profile accelerators for vector database search. IKS leverages CXL.cache to implement a seamless and efficient interface between the CPU and near-memory accelerators. 
        
    \item We implemented an end-to-end accelerated RAG application using IKS. IKS accelerates \enns for a 512GB knowledge store by 13.4--27.9$\times$, leading to a 1.7--26.3$\times$ end-to-end inference speedup for representative RAG applications. 
\end{itemize}

\section{Background}
\label{sec:background}

\def\candidate{indexed embedding\xspace}
\subsection{Information Retrieval in RAG}
\label{sec:background:rag}



Recent advancements in RAG indicate superior outcomes when employing dense retrieval over other methods, for uni-modal~\cite{izacard-grave-2021-leveraging, izacard2021distilling, 10.5555/3495724.3496517} and multi-modal~\cite{gui-etal-2022-kat, 10.1145/3539618.3591629, 10.1145/3578337.3605137} scenarios. Consequently, our emphasis in this study centers on dense retrieval models, exploring their efficiency-related aspects. 

In the context of dense retrieval, a query encoder, denoted as $E_q$, and a document encoder, denoted as $E_d$, are trained to encode queries and documents, respectively, and map them into a high-dimensional vector space. The similarity score between a document\footnote{The term ``document'' refers to any retrievable item from the knowledge source.} $d$ and a query $q$ is calculated as $s_d = E_q(q) \cdot E_d(d)$, where $E_q(q) \in \mathbb{R}^{h}$, $E_d(d) \in \mathbb{R}^{h}$ and $h$ is the hidden dimension of query and document encoders. 
Then documents are sorted based on their similarity scores and top documents are retrieved~\cite{dpr}. In a real RAG implementation, in an offline phase, all the documents are encoded into embedding vectors. The embedding vectors are stored in a vector database for dense retrieval. In the paper, we refer to the encoded documents as \textit{embedding vectors} and the vectors generated by the retriever model as \textit{query vectors}. 

For dense retrieval, two distinct search algorithms are prevalent: Exact Nearest Neighbor Search (\enns) and Approximate Nearest Neighbor Search (\anns). 
\enns exhaustively computes the complete pairwise distance matrix between embedding and query vectors.
In \anns, however, strategies such as Product Quantization (PQ)~\cite{5432202}, Inverted File with Product Quantization (IVFPQ)~\cite{6248038}, and Hierarchical Navigable Small World (HNSW)~\cite{hnsw}, are employed to reduce the search space, seeking to trade off a small amount of search accuracy for higher search efficiency.

\subsection{Applications of RAG}

RAG has proven beneficial for various tasks in natural language processing~\cite{Li2022ASO, reml, Kim-2024-Retrieval}, including dialogue response generation~\cite{weston-etal-2018-retrieve, Wu_Wei_Huang_Wang_Li_Zhou_2019, cai-etal-2019-skeleton, 9982598, thulke2021efficient, Bonetta_2021, tian-etal-2019-learning, 10.1007/978-3-031-44693-1_2, shuster-etal-2021-retrieval-augmentation}, machine translation~\cite{Gu2017SearchEG, xu-etal-2020-boosting, he-etal-2021-fast, zhang-etal-2018-guiding}, grounded question answering~\cite{izacard-grave-2021-leveraging, 10.5555/3495724.3496517,erag, izacard2021distilling, qu-etal-2021-rocketqa, 10.1162/tacl_a_00530, kilt, ium, urag, 10.1145/3626772.3657923}, abstractive summarization~\cite{peng-etal-2019-text, parvez-etal-2021-retrieval-augmented}, code generation~\cite{10.5555/3327546.3327670, liu2021retrievalaugmented}, paraphrase generation~\cite{kazemnejad-etal-2020-paraphrase, su-etal-2021-keep}, and personalization~\cite{salemi2023lamp, rspg, rag-vs-lora-personalization, Kumar-2024-LongLaMP}. Additionally, RAG's application extends to multi-modal data tasks like caption generation from images, image generation, and visual question answering~\cite{10.1145/3549555.3549585, Fei_2021, ramos-etal-2023-retrieval, Chen2022ReImagenRT, gui-etal-2022-kat, 10.1145/3539618.3591629, chen-etal-2022-murag}.

It is noteworthy that commercial LLM systems employing RAG are typically proprietary, and as such, their implementations are not openly accessible. Nevertheless, insights into the implementation of these systems can be gleaned from open-source releases by research labs within commercial entities. We adhere to a methodology akin to the approach outlined by~\citet{izacard-grave-2021-leveraging} and \citet{10.5555/3495724.3496517}, both of which are contributions from Meta AI. Our implementations closely align with the depicted pipeline in Figure~\ref{fig:rag-overview}. Specifically, we employ a dense document retrieval model as the retriever and leverage a language model for answer generation, consistent with the aforementioned work. Additionally, for efficient vector search capabilities, we utilize the Faiss~\cite{Jegou-2017-Faiss} library, similar to the aforementioned works.

\section{Demystifying \re}
\label{sec:profiling}

\def\qa_app{open-domain question answering\xspace}
\def\qa{QA\xspace}


In this section, we profile the end-to-end execution of three representative long-form question-answering RAG applications and quantify both the execution time breakdown and the generation accuracy of RAG with different hardware and software configurations: \appt, where we use the T5-based Fusion-in-Decoder~\cite{t5,izacard-grave-2021-leveraging} as the generative model, as well as \appm, and \appl, where we use 4-bit-Quantized Llama-3-8B-Instruct and Llama-3-70B-Instruct~\cite{meta-2024-llama3} as the generative models, respectively. The knowledge source for all workloads is Wikipedia, and a trained BERT base (uncased) model is used to generate embedding vectors for documents. We assume 16-bit number format and test with various 
vector database sizes (corpus size) that store the embedding vectors. The documents themselves are stored in the CPU memory. 

In \appt, the documents are presented via the Fusion-in-Decoder approach, where documents are encoded by the encoder stage of a T5 model, and these encoded representations are combined for the decoder stage. In \appm and \appl, retrieved documents are presented as plaintext in the prompt. For more information about the methodology, see Section~\ref{sec:expr:method}. 

In the following subsections, we discuss both the accuracy of an end-to-end RAG system and the retrieval model on its own. \textit{Retrieval accuracy} is discussed in terms of recall, where \enns is considered to be perfect, and the recall score of an \anns algorithm is the proportion of relevant documents retrieved by both \enns and \anns algorithm compared to the total number of relevant documents retrieved by \enns. \textit{Generation accuracy} refers to how well an end-to-end RAG system answers questions. For details on the evaluation of generation accuracy, see Section~\ref{sec:expr:software}.

\subsection{Tuning \re Software Parameters}
\label{sec:profiling:softwaretuning}

Both the retrieval phase and generation phase of the RAG systems that we use offer support for batching of queries in order to improve data reuse and to amortize data movement overheads over several queries. However, batching is not always an option in practice, particularly in the case of latency-critical applications. We consider batch sizes of 1 for latency-critical uses across all applications and 16 for throughput-optimized applications. Batch size does not impact generation accuracy and only affects execution time. 
An important parameter in RAG is ``K'' or the \textit{number of documents} retrieved and fed to the language model for text generation. Increasing the document count significantly impacts the generation time. The computation required for transformer inference scales at least linearly with the input size~\cite{transformer}, and if we concatenate the retrieved documents, we face significant computation and memory overhead~\cite{pmlr-v201-duman-keles23a,zhu-2024-accelerating}. In particular, the memory required to store a key-value cache entry for a single token can be computed as follows: $n_{\text{layers}}\times n_{\text{KV-heads}} \times d_{\text{head}} \times n_{\text{bytes}} \times 2$, where $n_{\text{bytes}}$ refers to the size of the number format~\cite{lienhart}. For \appm with a 16-bit number format, this is $32\times 8 \times 128\times 2 \times 2 = 131$ kB per token. While exact token counts depend on the tokenization process, each document (for all applications) is 100 words long; for \appm and \appl, this averaged 127 tokens per document across our evaluation dataset.

\subsection{Examining Approximate Search for RAG}
\label{sec:profiling:anns} 

An important algorithmic consideration that can impact the inference time and generation accuracy of RAG is the choice of retrieval algorithm from the vector database, where we can use exact nearest neighbor search (\enns) or approximate nearest neighbor search (\anns). The particular algorithm used for retrieval is implemented by a data structure called an \textit{index}, which stores the embedding vectors computed offline, as described in Section~\ref{sec:background:rag}. For \enns, an index is a wrapper around an array of embedding vectors sequentially iterated over during the search, but for \anns, the index can be more complex. For example, HNSW stores embedding vectors in a graph-based data structure~\cite{hnsw}. 

To evaluate \anns, we use the state-of-the-art HNSW~\cite{hnsw} \anns algorithm, and fine-tune the \textit{M} and \textit{efConstruction} parameters to maximize retrieval accuracy while maintaining a reasonable graph, yielding an index with \textit{M} of 32 and \textit{efConstruction} of 128. From this, we evaluate two configurations, \anns-1 and \anns-2, which use different \textit{efSearch} parameters: 2048 and 10000. In the context of an end-to-end RAG system, the trade-off of generation accuracy and runtime was evaluated for this index for various choices of \textit{efSearch}. A lower \textit{efSearch} provides higher search throughput, but lower generation accuracy, and a higher \textit{efSearch} provides lower search throughput, but higher generation accuracy. Other HNSW and IVFPQ indexes were tested but provided lower generation accuracy, or similar runtime to \enns (or even worse, in some cases), negating the benefits of approximation.  

\begin{figure}
    \centering

    \includegraphics[width=0.42\textwidth]{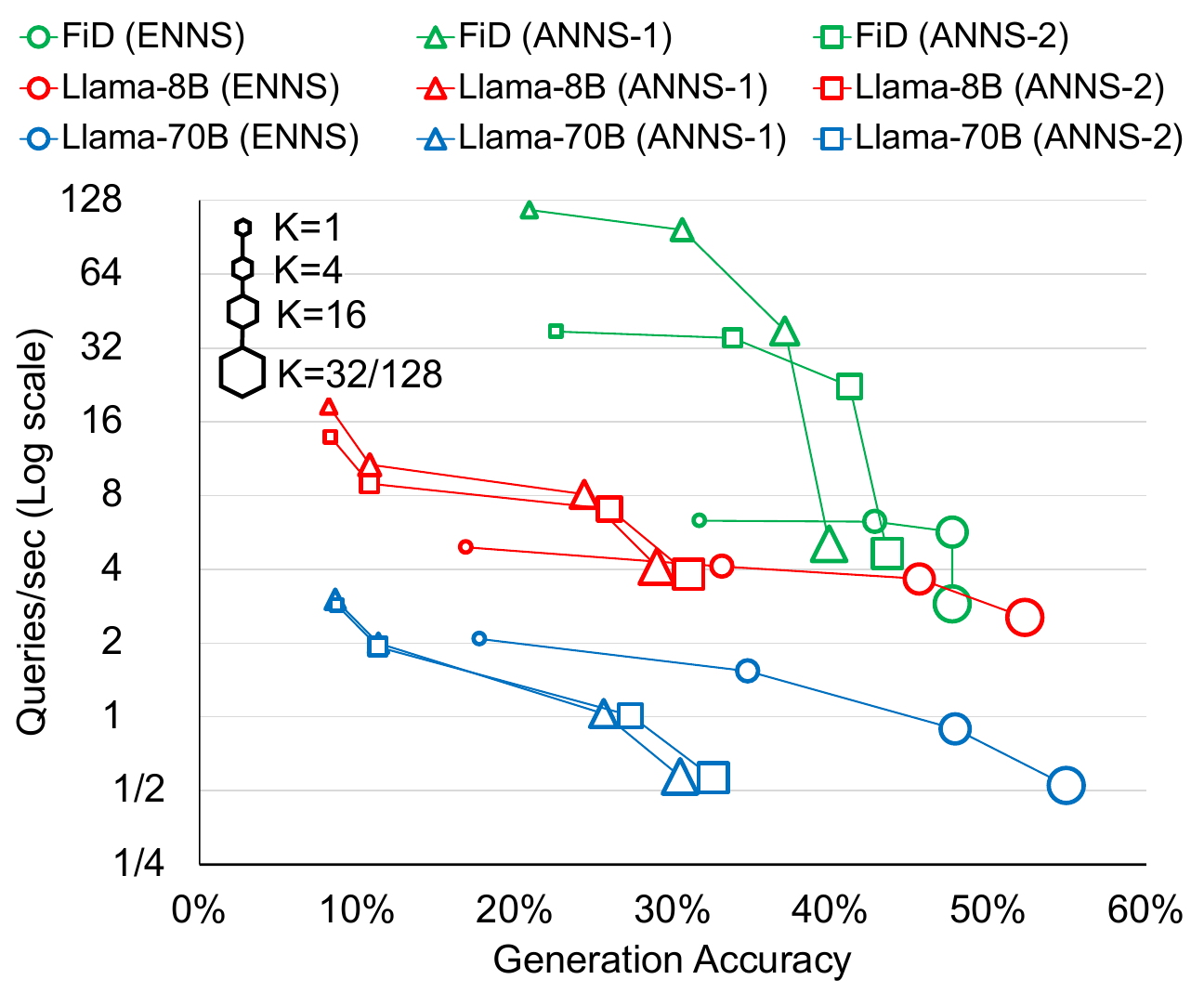}

    \caption{Generation accuracy vs. throughput (Queries/sec) of representative RAG applications for various retrieval algorithms and document counts (K). The corpus size is set to 50~GB and batch size to 16.}
    \label{fig:acc-tpt-demystifying}
    
\end{figure}

\noindent\textbf{Generation Accuracy with \anns vs. \enns:} Figure~\ref{fig:acc-tpt-demystifying} compares the generation accuracy and throughput of \anns- and \enns-based RAG applications for \appt, \appm, and \appl. The figure illustrates that retrieval quality strongly influences the end-to-end generation accuracy. 
As shown in Figure~\ref{fig:acc-tpt-demystifying}, with document count of one, compared to \enns, the generation accuracy of \anns-1 and \anns-2 drops by 22.6 and 34.0\% for \appt, 52.8 and 53.4\% for \appm, and 51.0 and 51.5\% for \appl, respectively. 
With a document count of 16, a similar trend in generation accuracy is observed, with \anns-1 and \anns-2 leading to an accuracy reduction of 13.6 and 22\% for \appt, 38.4 and 42.2\% for \appm, and 38.4 and 45.2\% for \appl, respectively. Interestingly, the impact of retrieval quality on generation accuracy appears to be even larger when using large models that have not been fine-tuned for this task.

Several prior works~\cite{ann-benchmarks,vdm_in_rd} demonstrate that hyper-parameter tuning can enhance the retrieval accuracy of \anns, potentially matching that of \enns across various workloads. While we optimized our HNSW indexes for accuracy and throughput, these indexes could not match \enns in end-to-end generation accuracy while achieving significantly (more than $2\times$) faster search. By using a small \textit{efSearch} value, retrieval speed improves significantly, allowing for the use of a larger value of K to compensate for the reduced retrieval quality. However, trading retrieval quality for retrieval speed in this way resulted in lower generation accuracy and end-to-end throughput compared to a larger \textit{efSearch}, where a higher-quality, slower search scheme permits greater accuracy at lower K values (thus reducing generation times). For example, \anns-2 with 16 documents have 3\% higher accuracy and 128\% higher throughput compared to \anns-1 with 128 documents for \appt. Further improving retrieval quality via exact search gives \enns-based RAG Pareto-superiority above sufficiently high accuracy thresholds ($\sim43\%$, $\sim27\%$, and $\sim14\%$ for \appt, \appm, and \appl, respectively) as demonstrated in Figure~\ref{fig:acc-tpt-demystifying}. In general, our findings highlight the potential for reducing generation time by leveraging high-quality retrieval methods when high accuracy is required.

\noindent\textbf{Scaling of \anns and \enns:} Previous works~\cite{hnsw,ann-benchmarks} identified the trade-off between retrieval quality and runtime, and challenges with high-quality \anns have motivated accelerators such as ANNA~\cite{anna} and NDSearch~\cite{ndsearch}. While lower-quality \anns algorithms could possibly provide orders of magnitude faster nearest neighbor search compared with \enns, high-quality \anns algorithms are shown to provide only a modest speedup~\cite{ANN_HDD,comprehensive_ANN}. For example, \anns-2, which is the best performing \anns configuration in Figure~\ref{fig:acc-tpt-demystifying}, offers only a 2.5$\times$ speedup compared with \enns. In fact, all the Pareto frontier configurations that provide high generation accuracy in Figure~\ref{fig:acc-tpt-demystifying} are \enns. Therefore, in the rest of this section, we focus on understanding how to optimize and accelerate RAG applications with \enns.

\begin{figure}
    \centering

         \begin{subfigure}[b]{0.45\textwidth}
             \includegraphics[width=\textwidth]{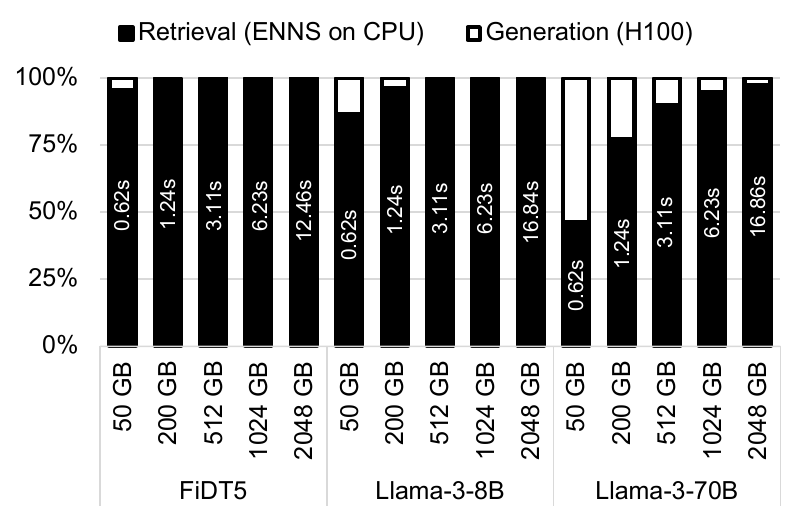}

    \caption{Sensitivity to corpus size. All configurations use K=16. 
    }
    \label{fig:latency:corpus}
         \centering

     \end{subfigure}

    \centering

         \begin{subfigure}[b]{0.45\textwidth}

    \includegraphics[width=\textwidth]{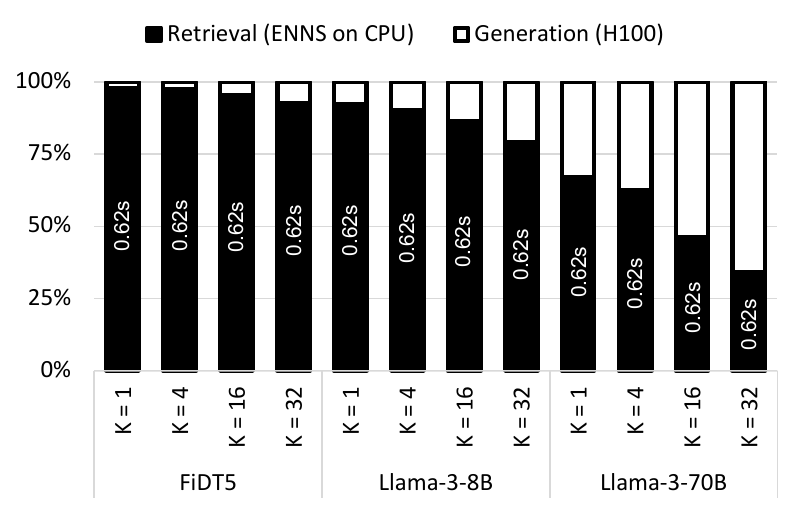}

    \caption{Sensitivity to K. All configurations use a 50 GB corpus.
    }
    \label{fig:latency:k}
     \end{subfigure}

     \caption{Latency breakdown of \appt, \appm, \appl for various values of K, corpus sizes. All configurations use batch size 1. Retrieval is \enns and runs on CPU, generation runs on a single NVIDIA H100 (SXM) for all generative models. \editlow{The value in each bar shows the absolute retrieval time.}}
    \label{fig:latency}

\end{figure}

\subsection{End-to-End RAG Performance with \enns}
\label{sec:profiling:accel-gen}

In this subsection, we profile time-to-interactive (also known as time to first token)~\cite{greener_llm} for the \appt, \appm, and \appl RAG applications and report latency ratios for the retrieval and generation phases. For all experiments, retrieval uses \enns and runs on the CPU, while generation runs on a single \editlow{NVIDIA H100 GPU}. We select CPU as the baseline for \enns retrieval, rather than GPU. \editlow{This decision is made based on the high cost of using GPU memory }

As we discussed in Section~\ref{sec:profiling:anns}, the generation accuracy of RAG applications directly depends on the retrieval accuracy. However, as shown in Figure~\ref{fig:latency:corpus}, utilizing \enns for retrieval can quickly become an end-to-end bottleneck in RAG applications, even for large models. 
Although it is possible to compensate for the retrieval accuracy by increasing K (in case of using \anns), as shown in Figure~\ref{fig:latency:k}, increasing K would increase the generation time and is costly in terms of time to first token.

The two phases in a RAG pipeline have different characteristics: \enns is extremely memory bandwidth-bound, and generation is relatively compute-bound. Nevertheless, the current state-of-the-art focus in building AI systems is only on accelerating the generation phase~\cite{piggyback,shen2023efficient, keller202395, xiao2023efficient, alizadeh2023llm, lin2023awq, xiao2023smoothquant, yang2023inference, miao2023specinfer, gim2023prompt,patel2024splitwise}. Next, we discuss the feasibility of accelerating high-quality nearest neighbor search for future RAG applications. 

\begin{table}
\smalltable
\centering
    \begin{tabular}{c|cccc}
        Batch Size & \multicolumn{2}{c}{1} & \multicolumn{2}{c}{16}\\
         Corpus Size& 50 GB  & 512 GB & 50 GB & 512 GB \\
        \hline 

         CPU& 1  & 1 & 1 & 1\\
         AMX& 1.05 & 1.02 & 1.10 & 1.09\\
         GPU & 5.2 & 36.9 & 6.0 & 43.7\\

    \end{tabular}
    \caption{Speedup of Intel AMX and GPU for \enns, relative to a CPU baseline. AMX speedup is flat for very small batch sizes, due to the memory-bound nature of similarity search. For 50GB and 512GB corpus size, 1 and 8 H100 GPUs are used, respectively.}
    \label{tab:retrieval_platforms}

\end{table}

\subsection{High-Quality Search Acceleration}

\label{sec:motivation:enns}

Given the sensitivity of RAG generation accuracy, latency, and throughput to the retrieval quality, it is imperative to focus exclusively on accelerating the retrieval phase of future RAG applications. In this subsection, we discuss the feasibility of accelerating high-quality \anns and \enns. 

\noindent\textbf{Acceleration of High-Quality \anns:} High-quality \anns can be as slow as \enns~\cite{ANN_HDD}. There are prior works aimed at building hardware accelerators for high quality \anns~\cite{anna,ndsearch} because GPUs are not effective at accelerating key \anns algorithms such as IVFPQ and HNSW~\cite{Jegou-2017-Faiss}. Unfortunately, the complex algorithms and memory access patterns used for \anns algorithms also make \anns accelerators highly task-specific; for example, ANNA~\cite{anna} and NDSearch~\cite{ndsearch} can only accelerate PQ-based and graph-based \anns algorithms, respectively. However, our experimental results, which are in line with prior findings~\cite{comprehensive_ANN}, show that different corpora are amenable to different \anns algorithms.

\noindent\textbf{Acceleration of \enns:} 
\enns can be accelerated using conventional SIMD processors such as GPUs and Intel AMX because the algorithm is simple and data-parallel. Table~\ref{tab:retrieval_platforms} compares the speedup of AMX and GPU against a CPU baseline. Although GPUs can significantly speed up \enns, as the corpus size increases, the cost of offloading \enns to GPUs increases significantly. For example, to fit the 50~GB and 512~GB corpus sizes tested in Table~\ref{tab:retrieval_platforms}, we need to use \editlow{1 and 8 H100 GPUs}, respectively. One of the key contributors to the cost of GPUs is the high-bandwidth memory (HBM) used to implement GPU main memory, which is several times more expensive than DDR or LPDDR-based memories~\cite{hbm_price}.  
Lastly, GPUs provision huge amounts of compute relative to memory bandwidth\footnote{NVIDIA H100 80GB provisions ~296 Flops/Byte and ~592 Int8 Ops/Byte}, meaning that a large GPU die is poorly utilized by the primarily memory-bound workload of \enns~\cite{ibrahim2021analyzing}.

\subsection{Summary}
\label{sec:profiling:summary}

The analysis presented in this section, using various software and system configurations for RAG applications, led to the following takeaways:

\begin{itemize}
    \item Generation accuracy, time to interactive, and throughput of RAG applications can be improved by using a slower but higher-quality retrieval scheme.
    \item When high-quality retrieval is used, the retrieval phase accounts for a significant portion of end-to-end runtime, regardless of whether the search is performed via \enns or high-quality \anns.
    \item Using GPUs to accelerate \enns is expensive, and GPUs are not able to accelerate high-quality \anns effectively or affordably.
    \item New accelerators for \anns are highly complex and task-specific due to the unique requirements of \anns algorithms, while \enns relies on a very simple scheme, making \enns simpler to accelerate than \anns.
\end{itemize}

\section{Case for Near-Memory \enns Acceleration}
\label{sec:idea}

\begin{figure}
    \centering
    \includegraphics[trim={0 0 0 0.5cm},width=\linewidth]{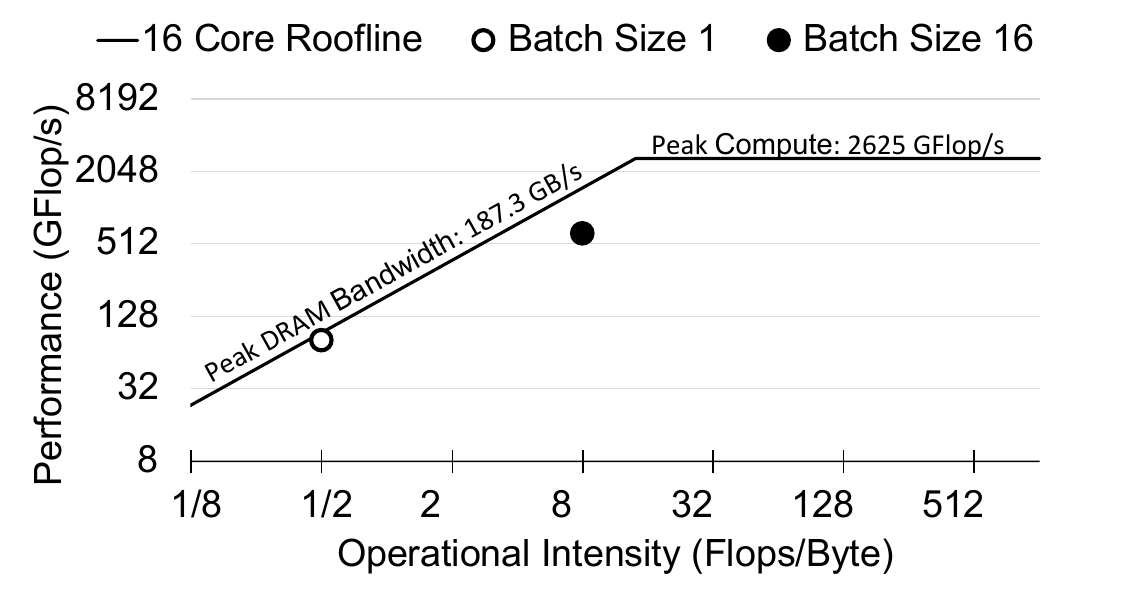}
    \caption{Roofline model for \enns using Batch Size 1 and 16. See Section~\ref{sec:expr:method} for the experimental setup.  
    } 
    \label{fig:roofline}
        \vspace{-10pt}
\end{figure}

\enns is characterized by the following features:

\begin{itemize}

    \item \enns operations exhibit no data reuse for pair-wise similarity score calculations between corpus vectors and a query vector.
    
    \item \enns operations consist of simple vector-vector dot-products coupled with top-K logic.
    
    \item \enns has a regular and predictable memory access pattern.
    
    \item \enns is highly parallelizable, allowing the corpus to be distributed across different processors with a simple aggregation of top-K similarities at the end.

\end{itemize}

These features make \enns a prime candidate for near-memory acceleration due to the following reasons: (1) Deep cache hierarchies are not beneficial for \enns and can even cause slowdown due to the complex cache maintenance and coherency operations managed by the hardware. This is evident from the roofline model in Figure~\ref{fig:roofline} as \enns running on the CPU cannot saturate the available DRAM bandwidth. (2) The limited data reuse with huge data set size enables low overhead software-managed cache coherency implementation between the host CPU and near-memory accelerators. (3) The regular memory access pattern of \enns enables coarse-grain virtual to physical address translation on near-memory accelerators. (4) \enns operations can be efficiently offloaded to a distributed array of near-memory accelerators that each operate in parallel on a shard of corpus data with a low-overhead top-K aggregation phase at the end.

Leveraging these unique features, we design, implement, and evaluate Intelligent Knowledge Store (IKS), a memory expander with a scale-out near-memory acceleration architecture, uniquely designed to accelerate vector database search in future scalable RAG systems. IKS is designed with three requirements in mind: (1) The memory capacity of IKS should be cost-effective and scalable because the size of vector databases for RAG applications is several tens or hundreds of gigabytes and is likely to increase. (2) The near-memory accelerators should be managed in userspace as the cost of context switches and kernel overhead would reduce the benefits of offloads. (3) The near-memory accelerators and host CPU should implement a shared address space; otherwise, explicit data movements between the CPU and near-memory accelerator address spaces will negate the benefits of near-memory offloads; another issue that GPU acceleration of \enns suffers from. Moreover, a partitioned address space requires rewriting the entire vector database application, as \enns is just one operation we want to accelerate near the memory, while other data manipulation operations, such as updates, should be managed by the host CPU.

We designed IKS, a type-2 CXL memory expander/accelerator, to meet all these requirements. Our rationale for choosing CXL over DDR-based (or DIMM-based)~\cite{alian-2018-mcn,axdimm-aquabolt,zhou_dimm-link_2023,patel-xfm-2023} near-memory processing architecture is that DIMM-based near-memory processing (1) requires sophisticated mechanisms to share the address space between near-memory accelerators and the host~\cite{smartdimm}, (2) limits per-rank memory capacity and compromises the memory capacity of the host CPU when used as an accelerator, and (3) has limited compute and thermal capacity. Instead, IKS relies on asynchronous CXL.mem and CXL.cache protocols to safely share the address space and independently scale the local and far memory capacity of the host CPU, implement a low-overhead interface for offloading from the userspace, and eliminate the limitations on the compute or thermal capacity of the PCIe-attached IKS card. In Section~\ref{sec:arch}, we explain the architecture of IKS and its interface to the host CPU, and how we used it to accelerate end-to-end RAG applications.

\section{Intelligent Knowledge Store}
\label{sec:arch}

\subsection{Overview}
\label{sec:arch:overview}

\begin{figure}[!ht]
    \centering 
    
    \begin{subfigure}[b]{0.70\linewidth}
        \centering
        \includegraphics[width=\linewidth]{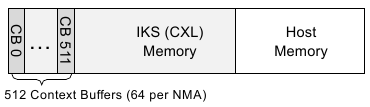}
        \caption{System Address Space with IKS}
        \label{fig:iks:addresspace}
    \end{subfigure}
    
    \begin{subfigure}[b]{0.7\linewidth}
        \centering
        \includegraphics[width=\linewidth]{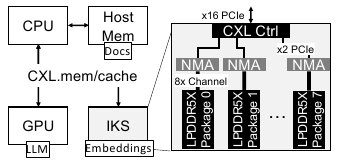}
        \caption{IKS system integration and architecture overview}
        \label{fig:iks:overview}
    \end{subfigure} 
    
    \begin{subfigure}[b]{0.8\linewidth}
        \centering
        \includegraphics[width=\linewidth]{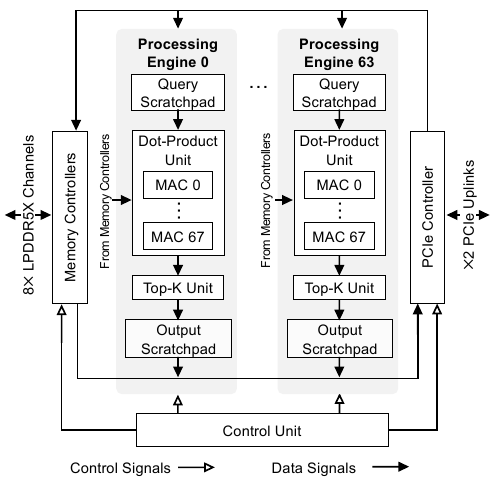}
        \caption{NMA internal architecture}
        \label{fig:iks:nma}
    \end{subfigure}
    
    \begin{subfigure}[b]{0.8\linewidth}
        \centering
        \includegraphics[width=\linewidth]{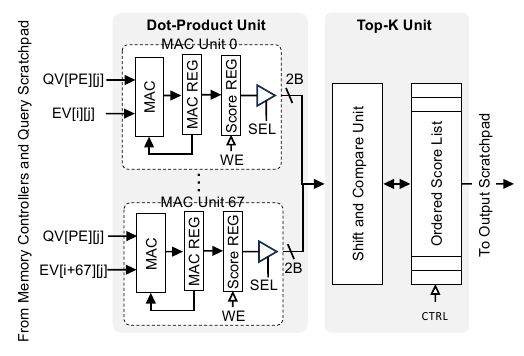}
        \caption{Dot-product unit and top-K units}
        \label{fig:iks:dotproduct}
    \end{subfigure}

    \caption{(a) IKS internal DRAM, scratchpad spaces, and configuration registers are mapped to the host address space. The scratchpad and configuration register address ranges are labeled as Context Buffers (CB). (b) IKS is a compute-enabled CXL memory expander that includes eight LPDDR5X packages with one near-memory accelerator (NMA) chip near each package. (c) Each NMA includes 64 processing engines. (d) Dot-product units reuse the query vector (QV) dimension across 68 MAC units. }
    \label{fig:iks}
\end{figure}

Figure~\ref{fig:iks} provides an overview of the Intelligent Knowledge Store (IKS) architecture. IKS incorporates a scale-out near-memory processing architecture with low-profile accelerators positioned near the memory controllers of LPDDR5X packages. While IKS can function as a regular memory expander, it is specifically designed to accelerate \enns over the embedding vectors stored in its LPDDR5X packages.

As shown in Figure~\ref{fig:iks:overview}, IKS utilizes eight LPDDR5X packages, each directly connected to a Near-Memory Accelerator (NMA) that implements both LPDDR5X memory controllers and accelerator logic. Each package contains 512Gb LPDDR5X DRAM with eight 16-bit channels, similar to CXL-PNM~\cite{park-2024-cxlpnm} and MTIA~\cite{mtia-meta}. One of the key differences between IKS and these architectures is the \textit{scale-out} near-memory acceleration architecture. IKS distributes the NMA logic over multiple chips, each providing high-bandwidth and low-energy access to its local LPDDR5X package.

\label{sec:arch:why_scale_out}
\noindent\textbf{Why Scale-Out NMA Architecture?} The rationale for such a scale-out NMA architecture is to keep the area of the NMA chip in check. Because memory PHYs are only implemented at the shoreline of a chip~\cite{loh-2015-interposer,mtia-meta,Patel-2024-CXL}, to implement 64 LPDDR5X memory channels, we need a chip with an approximate perimeter of 160~$mm$. This is because each LPDDR5X channel PHY approximately occupies a shoreline of 2.5~$mm$, based on the die shots of Apple M2~\cite{m2-die} in 5nm technology. A square-shaped chip with a 160~$mm$ perimeter has an area of 1600~$mm^{2}$, which is larger than the state-of-the-art lithography reticle limit~\cite{WikiChip-2024-Mask}. Although we can technically manufacture such a large accelerator using chiplets, the area of this huge multi-chip module would be wasted, as it is much larger than what is needed to implement the NMA logic, memory controllers, and PCIe/CXL controllers. For context, the area of an H100 GPU is 814~$mm^{2}$.

Splitting the NMAs into smaller chips increases the aggregate chip shoreline and improves yield. Using one NMA per LPDDR5X package requires only eight LPDDR5X memory channels per NMA, necessitating a minimum chip perimeter of 20~$mm$. IKS implements $\times$2 PCIe 5.0 to provide a 8~GBps uplink connecting each NMA to the CXL controller. With this design, the uplinks to the CXL controller are oversubscribed. Nevertheless, this oversubscription is neither a bottleneck for IKS operating in acceleration mode nor for IKS operating in memory expander mode. In acceleration mode, the bandwidth of local LPDDR5X channels is utilized for dot product calculations, and in memory expander mode, the data is interleaved over multiple LPDDR5X packages and read in parallel over the multiple $\times$2 PCIe uplinks.

\noindent\textbf{IKS is a type 2 CXL device.} IKS's internal memory is exposed as host-managed device memory where both the CPU and IKS can cache addresses within this unified address space (Figure~\ref{fig:iks:addresspace}). IKS leverages the low-latency accesses of CXL.mem and CXL.cache protocols to implement a novel interface between the near-memory accelerators and the CPU that: (1) eliminates the need for DMA setup and buffer management, and (2) eliminates the overhead of interrupt and polling for implementing notifications between the CPU and near-memory accelerators (\S\ref{sec:arch:cc-interface}).

\noindent\textbf{IKS supports spatial and coarse-grain temporal multi-tenancy.} In spatial multi-tenancy, the IKS driver partitions embedding vectors that belong to different vector databases across different packages, allowing each NMA to execute \enns independently per vector database. For temporal multi-tenancy, the IKS driver time-multiplexes similarity search in NMAs among different vector databases that store their embedding vectors in the same LPDDRX5 package. Time multiplexing takes place at the boundary of a complete similarity search. 

\noindent\textbf{Why LPDDR?} 
For IKS, a customized type-2 CXL device that should support cost-effective high capacity, neither HBM (expensive) nor DDR (general-purpose) are good options. LPDDR DRAM packages are integrated as part of system-on-chip designs, resulting in shorter interconnections, faster clocking, and less power wastage during data transmission. The most recent release of LPDDR, LPDDR5X, offers a bandwidth of 8533 Mbps per pin, exceeding that of DDR5, which provides a bandwidth of 7200 MTps. However, one challenge with using LPDDR in a datacenter setting is reliability, as LPDDR was originally designed for mobile systems. Although we could provision an in-line ECC processing block for error detection and correction, \enns similarity search is resilient to bit flips, and rare bit flips in \enns have negligible impact on the end-to-end RAG accuracy.

\subsection{Offload Model}
\label{sec:arch:offloadmodel}

The IKS address space is shared with the host CPU. 
The host CPU stores embedding vectors with a specific data layout (that we discuss in Section~\ref{sec:arch:datalayout}) in contiguous physical addresses in IKS, while the actual documents are stored in the host memory (either in DDR memory or CXL memory). The CPU runs the vector database application, which offloads the similarity calculations (i.e., dot-products between the query vectors and embedding vectors) using \texttt{iks\_search(query)}, a blocking API that does not require a system call or context switch. After each update operation, the vector database application will flush CPU caches to ensure that when \texttt{iks\_search(query)} is called, IKS does not contain any stale values.

\texttt{iks\_search(query)} hides the complexity of interacting with IKS hardware from the programmer by writing an \textit{offload context} to IKS and initiates an offload by writing into a doorbell register. \editlow{The offload context and doorbells are communicated through memory-mapped regions called \textit{context buffers} to the IKS as shown in Figure~\ref{fig:iks:addresspace}.} 
An \textit{offload context} includes query vectors, vector dimensions, and the base address of the first embedding vector stored in each LPDDR5X package. The host process then uses \texttt{umwait()} to block on the doorbell register (shared between IKS and the host and kept coherent via the CXL.cache protocol) to implement efficient notification between the paused CPU process and near-memory accelerators~\cite{yuan-rambda}. 

As IKS uses a scale-out near-memory processing architecture (\S\ref{sec:arch:overview}), the embedding vectors are distributed across different near-memory accelerators' local DRAM. Therefore, after all the near-memory accelerators complete the offload, the CPU process waiting on \texttt{umwait()} will be notified and execute an aggregation routine to construct a single top-K list. This top-K list is then used to retrieve the actual top-K documents from the host memory. \editlow{The CPU will locate documents based on the physical addresses of the top-K embedding vectors, as the addresses of the embedding vectors stored in IKS are known a priori.}

\subsection{Cache Coherent Interface}
\label{sec:arch:cc-interface}

\begin{figure}
    \centering
    \includegraphics[width=1\linewidth]{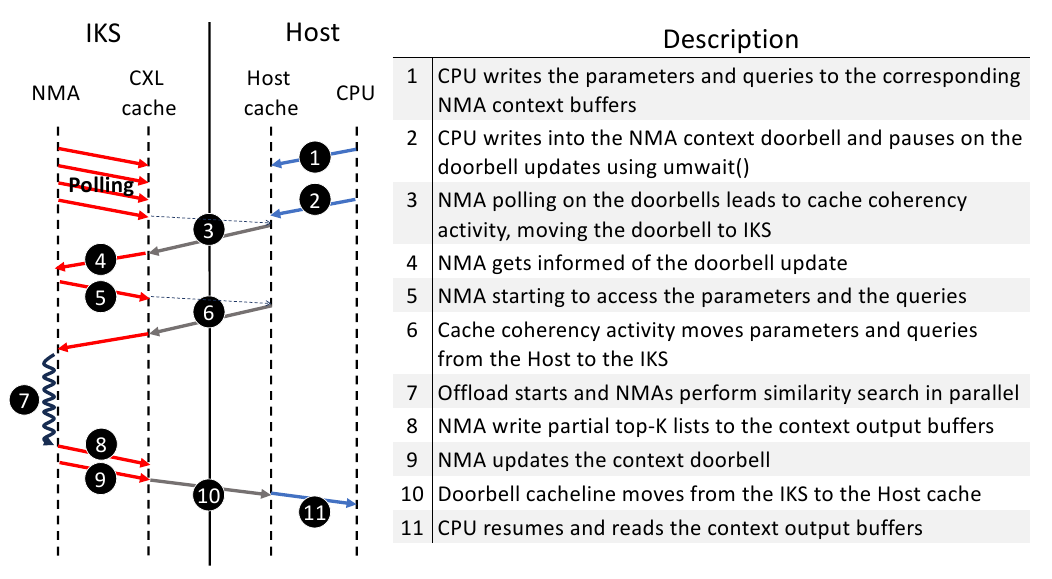}
    \caption{\editlow{CPU-IKS interface through cache coherent CXL interconnect.}}
    \label{fig:iks-trans}
    \vspace{-15pt}
\end{figure}

IKS leverages the cache-coherent interconnect in CXL.cache to implement an efficient interface between near-memory accelerators and host processes through shared memory. Figure~\ref{fig:iks-trans} illustrates the transactions through the CXL.cache interface between the host and IKS to initiate and conclude an offload. The host process writes the \textit{offload context} to the predefined \textit{context buffer} address range shared between NMAs and the host CPU (step 1). 
Note that the context buffer is cacheable, and the CPU uses temporal writes to populate the buffers. Next, the host process writes into a doorbell register, which is mapped to a cache line shared by NMAs. NMAs poll on the doorbell register, and as soon as there is a change, the offload starts (step 4). Once the host updates the doorbell register, it calls \texttt{umwait()} to monitor the register for changes from the IKS side.

Before computation in the NMA can start, the NMA reads the offload context from the IKS cache (step 5) and the context written by the host is moved to NMA's scratchpad. Once the NMA computation is complete, the NMA updates the context buffers with the partial list of similarity scores and physical addresses of the corresponding embedding vectors. Lastly, the NMA writes into the doorbell register, and the host gets notified of the completion of the offload through the \texttt{umwait()} mechanism (step 11).

Our experimental results on a two-socket Sapphire Rapids CPU show that communicating the offload context through cache-coherent shared memory provides 1.6$\times$ higher throughput compared with using non-temporal writes that mimic the PCIe MMIO datapath (i.e., CXL.io). Using a cache-coherent interconnect to implement the notification mechanism through the producer/consumer-style doorbell register eliminates the need for expensive interrupt or polling mechanisms.

\subsection{NMA Architecture}
\label{sec:arch:nma-arch}

As shown in Figure~\ref{fig:iks:nma}, each NMA implements 64 processing engines to accommodate similarity score calculations for up to 64 query vectors in parallel. Each processing engine includes a query scratchpad, dot-product unit, Top-K unit, and output scratchpad. There is a central control unit in each NMA that generates memory accesses, controls data movement within the NMA, and activates processing engines based on the number of query vectors provided by the host CPU. The network-on-chip implements a fixed broadcast network from DRAM to all the processing engines to reuse data when multiple processing engines are active and evaluate similarity scores against different query vectors.

As shown in Figure~\ref{fig:iks:dotproduct}, the dot-product unit includes 68 MAC units, each operating at a 1 GHz frequency and providing 68 GFLOPS (16-bit floating point multiply-accumulate operations) compute throughput; therefore saturating the 136 GBps memory bandwidth of the LPDDR5X channels. Each MAC unit evaluates the similarity score between the query (stored in the query scratchpad) and an embedding vector that is read from DRAM in \textit{VD} (Vector Dimension) cycles. All the processing engines operate on the same data that is read from the DRAM; in other words, each processing engine evaluates the similarity score between different query vectors and the same set of embedding vectors. Therefore, for a batch size of one, only one processing engine is utilized, and for a batch size of 64, all the processing engines are utilized. This way, we reuse the embedding vectors that are read from DRAM across different batch sizes.

As illustrated in Figure~\ref{fig:iks:dotproduct}, within an active dot-product unit, 68 MAC operations are performed in each clock cycle. The first input of the MAC units is dimension $j$ of the query vector in processing engine $PE$ (QV[PE][j]), and the second input is dimension $j$ of the embedding vectors $i$ to $i+67$ read from DRAM. As mentioned earlier, it takes \textit{VD} (Vector Dimension) cycles for a dot-product unit to evaluate the similarity score for a block of 68 embedding vectors. Once the similarity score is evaluated, it is loaded into a \textit{score register} (shown in Figure~\ref{fig:iks:dotproduct}) in the next clock cycle, and the MAC unit gets busy evaluating a new similarity score for the next 68 embedding vector block. The score registers (68 per processing engine) are then streamed out to the Top-K unit in the next 68 clock cycles.

The Top-K unit maintains an ordered list of the scores by comparing the incoming similarity scores with the head of the ordered list. Figure~\ref{fig:iks:dotproduct} illustrates the Top-K unit. If the value of the incoming score is larger, it is ignored; otherwise, it is inserted into the ordered list. Because the vector dimensions are much larger than 68, the serialized insertion into the ordered list is overlapped with the similarity score evaluations and is not on the critical path of the NMA offload.

After all the embedding vectors stored in the DRAM are evaluated, the control unit signals the end of the offload by loading the ordered Top-K list into the output scratchpad and writing to the doorbell register. The host CPU is then notified and can read the content of the output scratchpads through the CXL.cache protocol. Note that both the query scratchpads and the output scratchpads are mapped to the host memory address space. In the current incarnation of the NMA, the size of the query scratchpad (per processing engine) is 2KB, and we keep an ordered list of 32 scores (i.e., we set K to 32 in the hardware).

\begin{figure}
    \centering
    \includegraphics[width=0.47\textwidth]{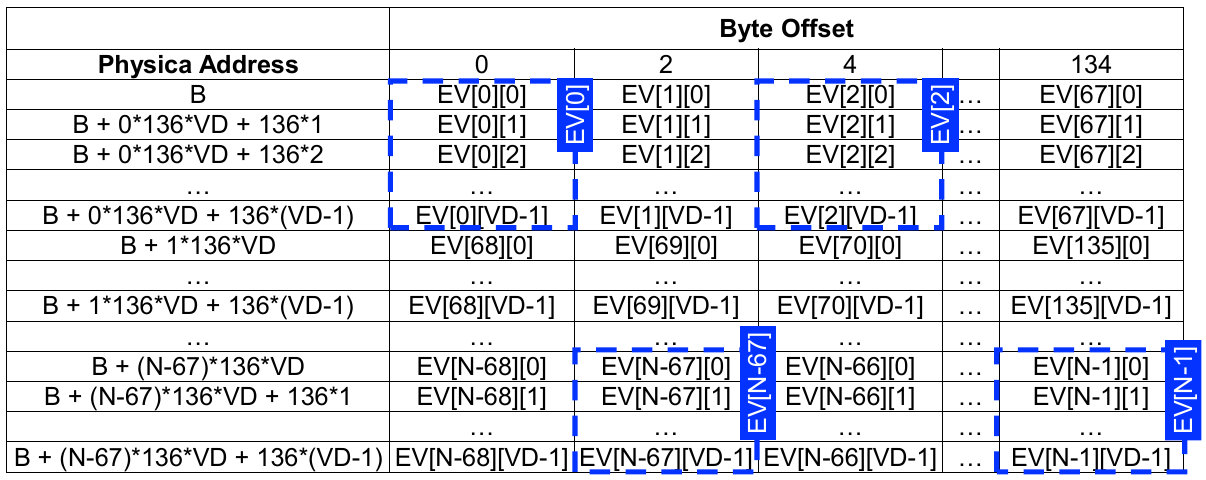}
    \caption{Data layout inside each LPDDR5X package. The host CPU communicates the base address "\textit{B}", vector dimension "\textit{VD}", and the number of vectors "\textit{N}" to the NMAs for each offload. Four embedding vectors (EVs) are highlighted in this layout.}
    \label{fig:datalayout-dram}
    \vspace{-15pt}
\end{figure}

\begin{figure}
    \centering
    \includegraphics[width=0.47\textwidth]{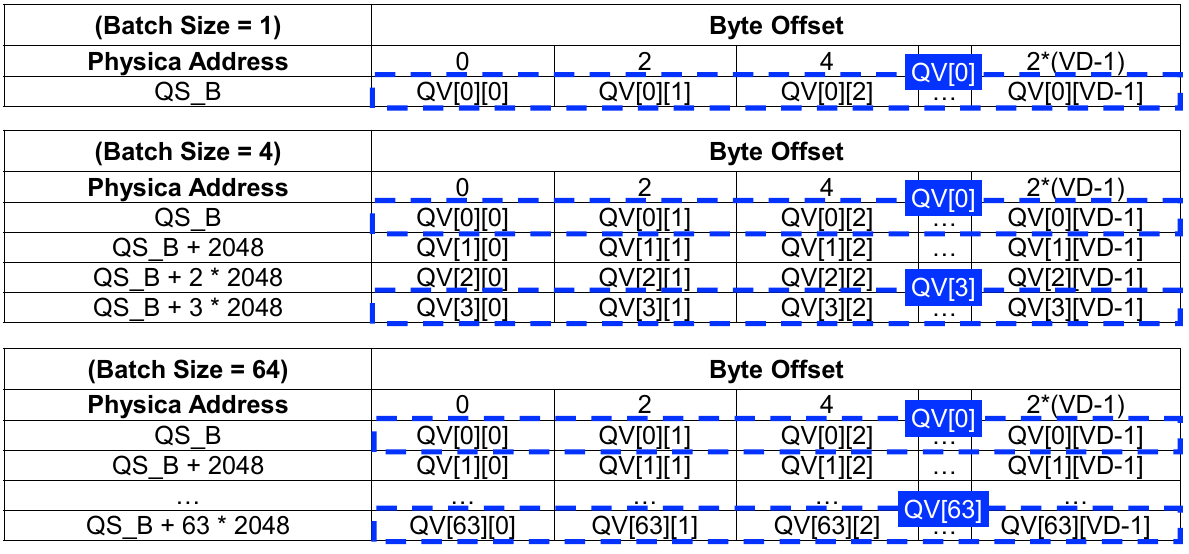}
    \caption{Data layout inside the query scratchpads mapped to host memory address at query scratchpad base address "QS\_B". As we increase the batch size, more query scratchpads are populated with distinct query vectors.}
    \label{fig:datalayout-sp}
    \vspace{-15pt}
\end{figure}

\subsection{Data Layout Inside DRAM and Query Scratchpad}
\label{sec:arch:datalayout}

The host CPU is required to store the embedding vectors in blocks of 68 vectors, laid out in the DRAM as shown in Figure~\ref{fig:datalayout-dram}. Because each embedding vector dimension is 2 bytes (16-bit floating point), each block is stored in $136 \times VD$ bytes within DRAM, where \textit{VD} is the vector dimension. Within a block, the embedding vectors are stored in column-major order. \editlow{This layout allows for efficient batching of corpus vectors, as each may be read and processed dimension-by-dimension.} Consequently, each NMA will access up to 136 bytes per cycle from the memory controller read queue, comprising one element from 68 distinct embedding vectors.

As discussed in Section~\ref{sec:arch:cc-interface}, the host CPU will fill the query scratchpads with query vectors before an offload starts. The query vectors are stored in sequential addresses within the query scratchpads, as illustrated in Figure~\ref{fig:datalayout-sp}.

This data layout inside DRAM and query scratchpads simplifies the address generation as well as the network-on-chip architecture of the NMAs. We modified the memory allocation scheme in the vector database application to implement the block data mapping of embedding vectors inside IKS DRAM as shown in Figure~\ref{fig:datalayout-dram}.

\section{Experimental Methodology}
\label{sec:expr:method}

\begin{table}[]
\smalltable
\vspace{0.5cm} 
    \centering

    \begin{tabular}{c|c|c}
        Platform & Parameter & Description\\
        \hline
        \multirow{ 8}{*}{CPU}&CPU model & Intel Xeon 4416+ 16 cores @ 2.00 GHz  \\
        &L1 Cache & 48 kB dcache, 32kB icache\\
        &L2 Cache & 2MB \\
        &L3 Cache & 37.5 MB shared \\
        &AVX& 2x AVX-512 FMA units (164 GFlop/s/core)\\
        &OS & Ubuntu 22.04.3 \\
        &Kernel & Linux 5.15.0-88-generic\\
        &Memory & 512 GB DDR5-4000 across 8 channels (256 GB/s) \\
        \hline
        AMX & -- & Intel AMX (BFloat16, 500 GFlop/s/core)\\

        \hline
        IKS (emulated)&  -- & 1.1 TB/s, 69.9 TFlop/s\\
        \hline
        GPU & GPU Model & \editlow{NVIDIA H100 SXM}: 3.35 TB/s, 1979 TFlop/s \\

    \end{tabular}
        \caption{Processing Element Options. \editlow{Memory configuration for Intel AMX is the same as for CPU.} }

    \label{tab:compute}
    \vspace{-15pt}

\end{table}

\subsection{Experimental Setup}

To evaluate the performance of the IKS, we developed a simulator (see appendix ~\ref{sec:appendix}) 
and fed \enns traces into it to obtain the retrieval time of IKS. The simulator is a cycle-approximate performance model that utilizes timing parameters from the RTL synthesis, LPDDR5X access timing, PCIe/CXL timing~\cite{li-2023-pond,schuh-2024-cc-nic}, along with calculations of real software stack overhead (top-K aggregation and \texttt{umwait()} overhead). It emulates an IKS as a CXL device running on a remote CPU socket. We implemented the end-to-end RAG application described in Section~\ref{sec:profiling} (i.e., \appt, \appm, and \appl), including the APIs for distributing queries to the NMA query scratchpad and reducing partial top-32 lists on the CPU. We ran the experiments on two servers equipped with Intel Xeon 4$^{th}$ generation CPUs and one NVIDIA H100 GPU NVIDIA GPUs. The system configuration is shown in Table~\ref{tab:compute}.

We implemented the RTL design of the Near-Memory Accelerator (NMA) used in IKS and synthesized it using Synopsys Design Compiler targeting TSMC's 16nm technology node. This process involved collecting key metrics such as area, power, and timing to ensure the design meets the optimal criteria for operation at 1 GHz. For other components, we estimated the area of the memory controllers and PHYs based on die shots from the Apple M2 chip, which utilizes LPDDR5 in a 5nm process~\cite{Locuza-2022-Die}. Since the area scaling of mixed-signal components is negligible~\cite{su-2017-amd,horowitz-2014}, we assumed the same area for the LPDDR5X PHYs and memory controllers when scaling to 16nm technology.


We developed a power model by evaluating the energy consumption of processing operations at the RTL level and incorporating the energy required for data access to scratchpads and LPDDR memory. For example, accessing data in SRAM consumes 39 fJ per bit, while LPDDR memory access requires 4 pJ per bit~\cite{domain-specific-hw-acc}. Since these energy values depend on the underlying technology node, we scaled them to correspond to a 16nm technology node for consistency~\cite{deepscale}.


\subsection{Software configuration}
\label{sec:expr:software}

Google's Natural Questions (NQ) dataset~\cite{natq,latent-retrieval} is used for the evaluation of models. Meta's KILT benchmark~\cite{Petroni-2020-KILT} divides these into training (\train) and validation (\dev) datasets. For the retrieval phase, we use a BERT base (uncased) model trained to perform similarity searches between questions and their supporting documents in \train. The document corpus is constructed as described in~\cite{dpr}, and an index is created using Faiss~\cite{Jegou-2017-Faiss} to perform the similarity search\footnote{We adapt the Faiss implementation of \enns by using Intel MKL as the BLAS backend, using BLAS for all batch sizes, and increasing the corpus block size from 1024 to 16384. See appendix \ref{sec:appendix}.}. Across \enns and \anns, Faiss is used for index management. The only change made in our evaluation is the use of Intel's OneMKL BLAS backend for \enns for all batch sizes, as this provided better performance than the default Faiss search scheme, which uses only BLAS for batch sizes 20 and above.

\noindent\textbf{\appt Application:} For testing the accuracy of \appt, as described in~\cite{izacard-grave-2021-leveraging}, the generator is initialized as a pretrained T5-base model (220 million parameters), then fine-tuned to predict answers from question-evidence pairs in the \train dataset.

To evaluate \appt on the \dev dataset, we use the exact match metric~\cite{rajpurkar2016squad}, which normalizes answers and compares them against a list of acceptable answers. For \appt, \textit{generation accuracy} scores refer to the percentage of \dev questions for which the RAG application generates a correct answer based on this exact match criterion.

\noindent\textbf{\appm and \appl Applications: }
To evaluate \appm and \appl on the \dev dataset, we guide the model via prompting and evaluate \textit{generation accuracy} using a Rouge-L ``recall'' metric~\cite{rouge}, which scores answer predictions based on the proportion of the correct answer that is continuously present in the predicted answer. The model is instructed to give a short answer and to answer only if it is ``completely sure.'' The prompting approach is used over fine-tuning to reflect an implementation that preserves the generality of the models. However, the downside of this approach is that evaluation is limited by prompt adherence, which is why the ``recall'' metric is used over precision or F1-Score. When evaluating end-to-end RAG systems, the applications process a batch of queries by first performing retrieval, then generation, before processing the next batch.

\section{Experimental Results}
\label{sec:expr}

\subsection{Effectiveness \edit{and Scalability} of IKS Retrieval}
\label{sec:exp:effectiveness}

\begin{figure}
    \centering
    \includegraphics[clip,width=0.49\textwidth]{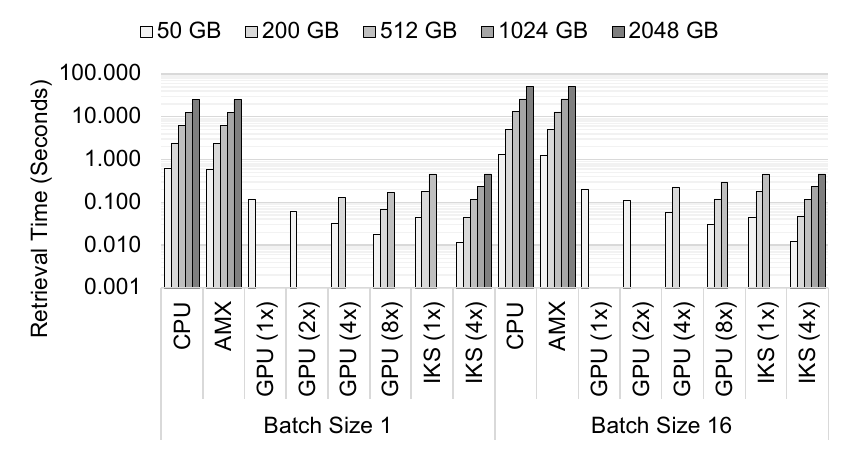}
    \caption{\edit{Comparison of \enns retrieval time for CPU, AMX, GPU (1, 2, 4, and 8 devices), and IKS (1, and 4 devices) for various corpus sizes. The absence of bars in specific GPU and IKS configurations indicates that the corpus exceeds the capacity of the accelerator memory. The Y-axis is in log-scale. }}

    \label{fig:scalability}
\end{figure}

\begin{figure*}
     \centering
     \begin{subfigure}[b]{0.32\textwidth}
         \centering
         \includegraphics[width=\textwidth]{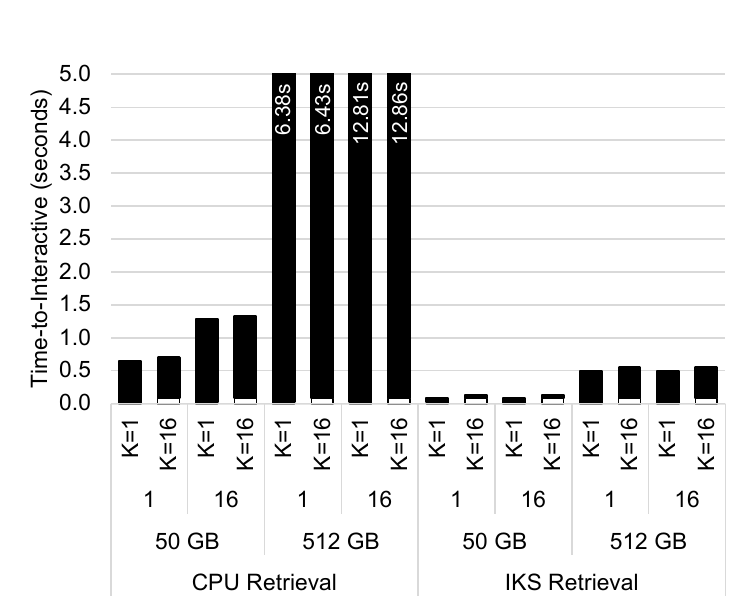}
         \caption{\appt}
         \label{fig:end-to-end:fid}
     \end{subfigure}
     \begin{subfigure}[b]{0.32\textwidth}
         \centering

         \includegraphics[width=\textwidth]{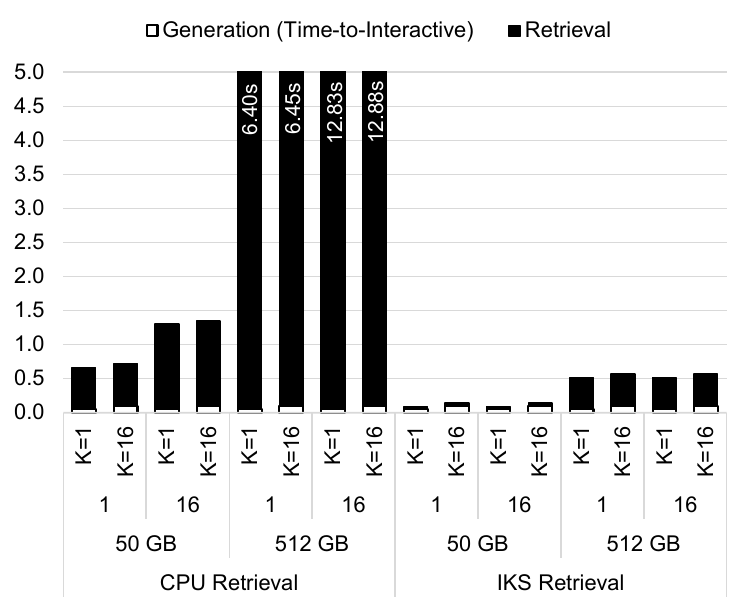}
        
         \caption{\appm}
         \label{fig:end-to-end:mistral}
     \end{subfigure}
          \begin{subfigure}[b]{0.32\textwidth}
         \centering
         \includegraphics[width=\textwidth]{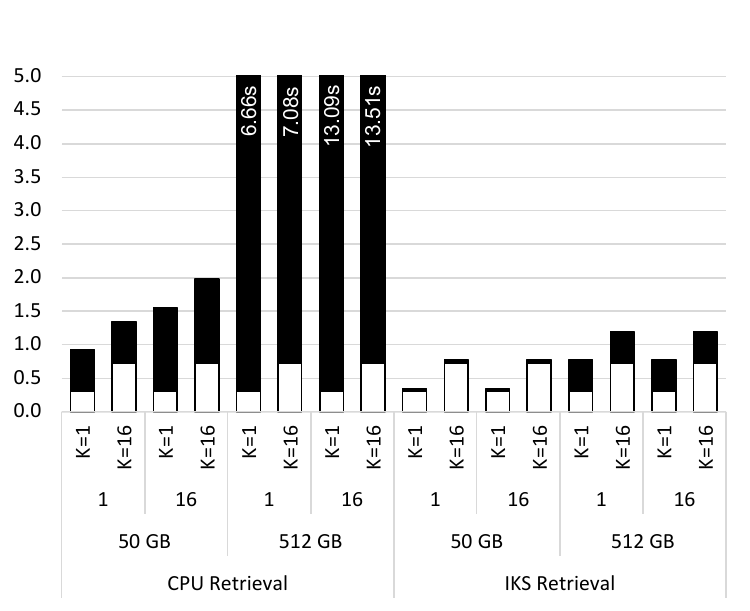}
         \caption{\appl}
         \label{fig:end-to-end:lamma}
     \end{subfigure}
          \vspace{-5pt}

        \caption{ Inference time breakdown of CPU vs. IKS retrieval for \appt, \appm, and \appl. Generative model runs on GPU. }

        \label{fig:end-to-end}
        \vspace{-10pt}
\end{figure*}

\edit{Figure~\ref{fig:scalability}} compares the performance of IKS with CPU, AMX (idealized, based on speedup for matrix multiplication), and GPU \enns retrieval. IKS provisions compute and memory bandwidth to balance the pipeline at the maximum batch size of 64; as such, performance is almost flat for batch sizes less than 64. As shown, the purposefully built NMA logic for \enns enables 1 IKS unit to outperform 1 GPU for a 50 GB corpus for batch sizes 1 and 16 by 2.6$\times$ and 4.6$\times$, respectively. \edit{This counterintuitive speedup of IKS over GPUs, which theoretically have both higher FLOPS and memory bandwidth than IKS, is due to two reasons: (1) top-K tracking and aggregation on GPUs is not efficient, while IKS includes specialized Top-K units; and (2) low utilization of the GPU chip translates to limited memory bandwidth usage, as saturating the entire HBM memory bandwidth requires many streaming multiprocessors and tensor cores to issue memory accesses to DRAM in parallel.} 

To demonstrate the scalability of IKS, we include the retrieval time of multi-GPU and multi-IKS setups. Because each H100 GPU can fit 80 GB of embedding vectors, 8 GPUs can accommodate maximum corpus size of 640 GB. However, with only four IKS devices, we can fit up to a 2 TB corpus size. As shown in Figure~\ref{fig:scalability}, with additional GPUs and IKS units, the retrieval time for the same corpus size decreases, demonstrating the high data-level parallelism of \enns and the strong scaling of both GPU and IKS. For example, GPU retrieval time for a 50 GB corpus size reduces by 1.9$\times$, 3.6$\times$, and 6.9$\times$ with 2, 4, and 8 GPU devices, respectively, and IKS retrieval time for a 50 GB corpus size reduces by 1$\times$ and 3.9$\times$ with 1 and 4 IKS units, respectively. Due to the low-overhead IKS-CPU interface, the dominance of similarity search latency in end-to-end \enns retrieval, and the highly parallelizable nature of \enns, IKS also provides near-perfect weak scaling. For instance, the retrieval time for a 2 TB corpus on 4 IKS units
is only 100$\mu$s longer than for a 512 GB corpus on 1 IKS unit. However, we do not evaluate configurations with more than four IKS units, and the overhead of host-side final top-K aggregation scales as additional units are added. Additionally, we do not evaluate deployments of IKS spanning multiple nodes.

\begin{table}[bp]
\smalltable
    \centering
    \begin{tabular}{c|cccc}
        Corpus Size & \multicolumn{2}{c}{50 GB} & \multicolumn{2}{c}{512 GB} \\
         
         Batch Size &  1  &  64 & 1 & 64\\
         \hline
         Write Query Vector & 0.3 us & 1 us & 0.3 us & 1 us\\
         
         Dot-Product & 45.96 ms& 45.96 ms & 470.6ms & 470.6 ms \\
         
         Partial Top-32 Read & 0.7 us & 22.4 us & 0.7 us & 22.4 us \\
         Top-K Aggregation & 19 us & 540 us & 23 us & 390 us \\

         \hline
         Total & 46.0 ms & 46.5 ms & 470.6 ms & 471.0 ms \\
    \end{tabular}
    \caption{Breakdown of \enns latency on IKS.}
    \label{tab:IKS breakdown}
\end{table}

Table~\ref{tab:IKS breakdown} reports the absolute time breakdown of \enns retrieval on IKS. We break down the retrieval time of IKS into four components: transfer time of query vectors over the CXL interconnect to the NMAs, time for performing dot-products (both computation and DRAM accesses), updating the top-k score lists in parallel on all NMAs, and time for reducing the partial top-32 lists into a single one on the CPU. The retrieval time of IKS does not change with the value of K (with a maximum K value of 32). This is because IKS always returns 32 top similarity scores, and it is up to the retriever model to pass between 1 to 32 of them to the generative model. As shown in the table, the majority of time is spent on computations and DRAM accesses, and the overhead of initiating offload over the cache-coherent interconnect and aggregating top-K documents on the CPU is negligible.

\subsection{End-to-End Performance}

Figure~\ref{fig:end-to-end} compares the end-to-end inference time of \appt, \appm, and \appl when CPU and IKS are used for \enns retrieval for various batch sizes, document counts, and corpus sizes. As shown, for large corpus sizes or large batch sizes, the inference time of the RAG applications with CPU retrieval exceeds several seconds, which is not acceptable for user-facing question-answering applications. IKS significantly reduces the \enns retrieval time for the applications. The end-to-end inference time speedup provided by IKS ranges between 5.6 and 25.6$\times$ for \appt, between 5.0 and 24.6$\times$ for \appm, and between 1.7 and 16.8$\times$ for \appl for various batch sizes, corpus sizes, and document counts.

\begin{figure}
    \centering
    \includegraphics[clip,width=0.42\textwidth]{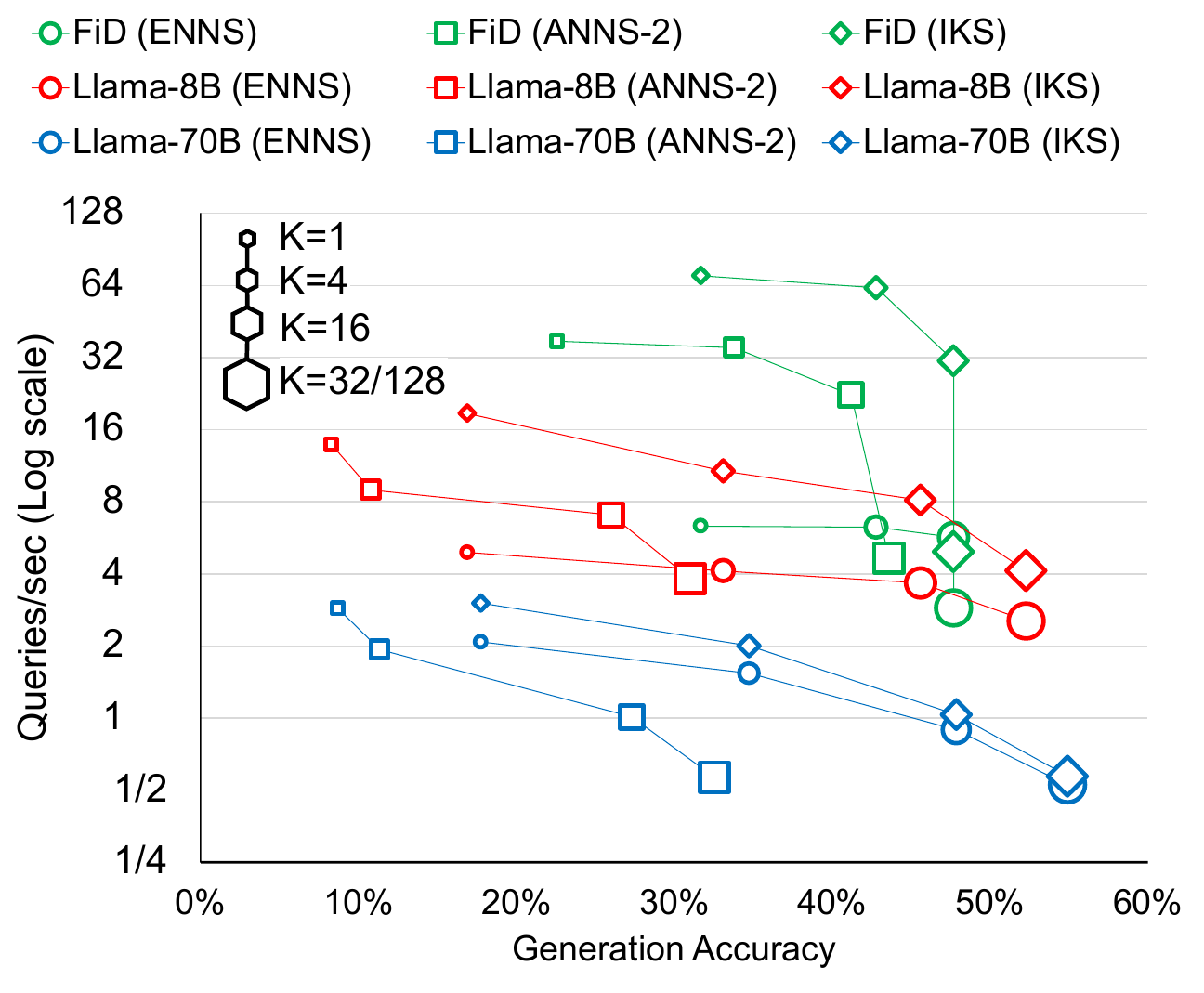}
    \caption{Comparison of accuracy and throughput of \appt, \appm, and \appl for various configurations. \anns-2 is an HNSW index with \textit{M}, \textit{efConstruction}, and \textit{efSearch} of 32, 128, and 2048, respectively. 
    }
        \vspace{-15pt}

    \label{fig:acc-tpt}
\end{figure}

To gain a comprehensive understanding of how the performance and accuracy of RAG applications with IKS acceleration compare across various configurations, Figure~\ref{fig:acc-tpt} depicts the queries per second and accuracy of \appt, \appm, and \appl implemented using four different configurations: RAG with \enns running on CPU, RAG with \anns (two configurations) running on CPU, and RAG with \enns running on IKS. The generative model runs on the GPU in all these configurations. As illustrated in Figure~\ref{fig:acc-tpt}, although \anns-2 configurations exhibit higher throughput compared to \enns (running on the CPU), their accuracy is lower. For RAG applications that use IKS, retrieval is not a bottleneck, and throughput is significantly improved, even compared to \anns, as the same generation accuracy can be achieved with smaller values of K (i.e., smaller but more accurate context sent to the generative model).

\subsection{Power and Area Analysis} 

The area of each NMA, which contains 64 processing engines, each comprising a dot-product unit, a 2 KB SRAM query scratchpad, a top-K unit, and an output scratchpad, is approximately 3.4~mm\(^2\) in the 16nm TSMC technology node. Additionally, 14~mm\(^2\) is required for the PHYs and memory controllers. However, the area of the NMA chip is determined by the shoreline because the 21~mm of shoreline required per NMA (20~mm for the LPDDR5X PHYs and 1~mm for PCIe PHYs \S\ref{sec:arch:overview}) necessitates that the NMA occupy at least 27.56~mm\(^2\) in the 16nm technology node. The NMA can be manufactured using older technology nodes to reduce costs and prevent area wastage, as the PHY area (which is mixed-signal) does not scale at the same rate as SRAM and logic~\cite{horowitz-2014,su-2017-amd}.


For a batch size of 1 and vector dimensions of 1024, the processing engines, along with the corresponding query scratchpad accesses, consume approximately 59~$mW$, while accessing embedding vectors from LPDDR memory requires 4.35~$W$. As a result, the total power consumption of IKS for a batch size of 1 is 35.2~$W$. With larger batch sizes, data reuse ensures that the power required for LPDDR access remains constant, but the power consumption of the processing engines increases linearly as more engines are activated to handle the additional workload. For instance, at full utilization with a batch size of 64, the total power consumption increases to 65~$W$.

\subsection{Cost and Power Comparison with GPU}
\label{sec:expr:gpucost}

IKS utilizes LPDDR5X memory to store embedding vectors. While figures for the cost of LPDDR5X are not yet available, we assume that HBM is more than $3\times$ more expensive than LPDDR~\cite{hbm_price}. Since a single IKS unit includes $6.4\times$ as much onboard memory as a single NVIDIA \editlow{H100 GPU}, the memory cost of IKS is expected to be approximately $2.5\times$ greater than that of a GPU.

For the comparison of compute unit cost, the GPU has a die area of 826 $mm^2$, while the IKS NMAs total a die area of 220 $mm^2$. Because the production cost of a chip increases superlinearly with die area~\cite{naffziger-2020-amdchiplet}, an IKS unit (with $5\times$ larger memory capacity) is expected to cost a fraction of a GPU.

\section{Discussion}

IKS provides a cost-effective solution for accelerating \enns, where the quality of the search is not dataset-dependent. However, if the dataset is amenable to clustering, then the accuracy gap between \enns and \anns would reduce, making \anns more attractive for retrieval. Moreover, IKS is best-suited to RAG applications requiring very high recall, and for datasets difficult to search with existing \anns schemes with relatively large batch sizes. For example, modern \anns schemes cannot eliminate more than 99\% of the search space for the GloVe dataset~\cite{comprehensive_ANN}, so at least 64\% of the corpus must be read by an \anns that does not offer data re-use across queries; in which case the overheads of common \anns schemes reduces performance to below that of \enns. However, for datasets that are easier to filter, there is an opportunity for improvement by incorporating approximation techniques into IKS; however, this introduces significant challenges as IKS owes much of its performance to the sequential memory access pattern of \enns.

One key inefficiency of IKS is that it performs an exhaustive search over the entire corpus, which consumes energy and saturates memory bandwidth. The high internal memory bandwidth utilization of \enns can cause slowdowns for external accesses by other applications that use IKS as a memory expander, rather than a vector database accelerator. Exploring early termination of similarity search~\cite{busolin-2024-early, Li-2020-improving} could be a natural solution for reducing the memory bandwidth utilization of \enns without compromising search accuracy.

Another inefficiency in the current version of IKS is the low NMA chip utilization for batch sizes less than 64. The rationale for overprovisioning NMA compute is that we effectively have free area on the NMA chip. Note that each NMA chip requires eight LPDDR5X memory channels, which demand 20~mm of chip shoreline. Therefore, the minimum NMA chip area is 25~mm$^{2}$ ($\S\ref{sec:arch:why_scale_out}$). Thus, the area on NMA is effectively free up to a cap of 25~mm$^{2}$. We chose to utilize this ``free'' area to overprovision compute so that IKS remains memory-bandwidth bound for all batch sizes below 64. There are opportunities for circuit-level techniques, such as clock and power gating, to power off extra processing engines when the batch size is below 64. Moreover, dynamic voltage and frequency scaling can be used to reduce the frequency and voltage of the NMA chip for batch sizes less than 64, allowing multiple processing engines to perform similarity searches for each query vector.
\section{Related Work}

\citet{cxl_accel_mem} implement a computational CXL memory solution for near-memory processing and showcased \enns acceleration inside the CXL memory. However, this work implements CXL memory using DDR DRAM, which does not meet the power and bandwidth requirements for \enns on large corpus sizes used in RAG. Additionally, our work implements a novel interface between host and near-memory accelerators through CXL.cache. \citet{anna} and \citet{ndsearch} present near-data accelerators for PQ- and Graph-based \anns, respectively. However, we accelerate \enns because different corpora are amenable to different \anns algorithms, and the complex algorithms and memory access patterns of such \anns schemes also make \anns accelerators highly task-specific. \citet{ke-2022-axdimm} propose near-memory acceleration of DLRM on Samsung AxDIMM. AxDIMM is based on a DIMM form factor that limits per-rank memory capacity and compromises the memory capacity of the host CPU when used as an accelerator (\S\ref{sec:idea}). In contrast, IKS does not strand the internal DRAM space and does not have capacity or compute throughput limitations.

\editlow{Concurrent with our work, others have also observed that low-quality retrieval can lead to both low-quality and slow generation. Corrective RAG filters out irrelevant documents from the retrieved list before sending them to the LLM~\cite{yan-2024-corrective}, while Sparse RAG enables LLMs to use only highly relevant retrieved information~\cite{zhu-2024-accelerating}. In this work, we used \enns to eliminate the risk of low-quality retrieval and reduce the context size. }

\section{Conclusion}
\label{sec:conclusion}

In this work, we profiled representative RAG applications and showed that the retrieval phase can be an accuracy, latency, and throughput bottleneck, highlighting the importance of an exact, yet high-performance and scalable retrieval scheme for future RAG applications. We designed, implemented, and evaluated the Intelligent Knowledge Store (IKS), a CXL-type-2 device for near-memory acceleration of exact K nearest neighbor search. The key novelty of IKS is the hardware/software co-design that enables a scale-out near-memory processing architecture by leveraging cache-coherent shared memory between the CPU and near-memory accelerators. IKS offers \editlow{18-52$\times$} faster exact nearest neighbor search over a 512 GB vector database compared to executing the search on Intel Sapphire Rapids accelerators, leading to \editlow{2.0-49$\times$} lower end-to-end RAG inference time.


\begin{acks}
This work was supported in part by NSF grant numbers 2239020, 1565570, and 2402873, in part by ACE, one of the seven centers in JUMP 2.0, a Semiconductor Research Corporation (SRC) program sponsored by DARPA, in part by the Office of Naval Research contract number N000142412612, and in part by the Center for Intelligent Information Retrieval. Any opinions, findings, conclusions, and recommendations expressed in this material are those of the authors and do not necessarily reflect those of the sponsors. We thank Jae-sun Seo and Yuan Liao from Cornell University for their help in synthesizing the near-memory accelerators on 16nm TSMC technology.  
\end{acks} 

\appendix
\section{Artifact Appendix}
\label{sec:appendix}

\subsection{Abstract}
This appendix describes two artifacts: 1–The cycle-approximate simulator for IKS, which models IKS using timing data gathered from RTL synthesis. 
2–FAISS modified for fast ENNS on Intel CPUs. All artifacts are available via Github.

\subsection{Artifact check-list}

\begin{itemize}
    \item \textbf{Simulator:} \href{https://github.com/architecture-research-group/iks_simulator}{https://github.com/architecture-research-group/iks\_simulator}
    \item \textbf{Optimized Faiss:} \href{https://github.com/architecture-research-group/ae-asplo25-iks-faiss/tree/main}{https://github.com/architecture-research-group/ae-asplo25-iks-faiss/tree/main}
   
    \item \textbf{Compilation:} Please refer to each program's repository.
    \item \textbf{OS requirement:} Modern Linux kernel
    \item \textbf{Hardware requirement:} Intel 4th Gen Xeon Scalable Processors or newer, with AMX equipped and enabled.
    \item \textbf{Software requirement:} Intel MKL Installed

    \item \textbf{Publicly available?:} Yes.
\end{itemize}

\balance
\bibliography{refs}

\end{document}